\documentclass[conference]{IEEEtran}
\usepackage{hyperref}
\usepackage{cite}
\usepackage{wrapfig}
\usepackage{amsmath,amssymb,amsfonts}
\usepackage{algorithmic}
\usepackage{graphicx}
\usepackage{textcomp}
\usepackage{xcolor}
\usepackage{multirow}
\usepackage{algorithm}

\def\BibTeX{{\rm B\kern-.05em{\sc i\kern-.025em b}\kern-.08em
    T\kern-.1667em\lower.7ex\hbox{E}\kern-.125emX}}

\newcommand{\wh}[1]{\widehat{#1}}
\newcommand{\wt}[1]{\widetilde{#1}}

\def\ba#1\ea{\begin{align}#1\end{align}}
\def\mkakko#1{\left(#1\right)}
\def\ckakko#1{\left\{#1\right\}}
\def\kkakko#1{\left[#1\right]}

\begin{document}
\title{Distributional Actor-Critic Ensemble for Uncertainty-Aware Continuous Control}

\author{\IEEEauthorblockN{Takuya Kanazawa, 
Haiyan Wang, Chetan Gupta}
\IEEEauthorblockA{\textit{Industrial AI Lab, Hitachi America, Ltd.~R\&D, Santa Clara, CA}
\\Email:~takuya.kanazawa@hal.hitachi.com
}}

\maketitle

\begin{abstract}
Uncertainty quantification is one of the central challenges for machine learning in real-world applications. In reinforcement learning, an agent confronts two kinds of uncertainty, called epistemic uncertainty and aleatoric uncertainty. Disentangling and evaluating these uncertainties simultaneously stands a chance of improving the agent’s final performance, accelerating training, and facilitating quality assurance after deployment. In this work, we propose an uncertainty-aware reinforcement learning algorithm for continuous control tasks that extends the Deep Deterministic Policy Gradient algorithm (DDPG). It exploits epistemic uncertainty to accelerate exploration and aleatoric uncertainty to learn a risk-sensitive policy. We conduct numerical experiments showing that our variant of DDPG outperforms vanilla DDPG without uncertainty estimation in benchmark tasks on robotic control and power-grid optimization.
\end{abstract}


\section{Introduction}

Nowadays Artificial Intelligence (AI) is becoming more and more popular as a tool to support human decision making in complex environments. Among others, Reinforcement Learning (RL) is a promising approach to tackle temporally extended hard optimization problems \cite{DBLP:books/lib/SuttonB98}. RL has already shown excellent performance in computer games \cite{DBLP:journals/nature/MnihKSRVBGRFOPB15} and even surpassed human champions in Go, Chess and Shogi \cite{
2018Sci...362.1140S}. Unlike traditional optimization methods, RL is capable of directly learning from high-dimensional sensory inputs with the aid of deep neural networks, which makes it an appealing method of choice for industrial tasks such as autonomous vehicle control and factory automation. However, RL shares several difficulties with other machine learning methods, which impedes immediate applications of RL to real-world problems. First, RL is quite data hungry, typically requiring hundreds of thousands of trial and error in simulations until a good policy is learned. Second, RL may get stuck in local optima and fail to find a globally optimal policy due to insufficient exploration of the environment. Third, RL may propose to take a wrong or even harmful action when it confronts out-of-distribution data that was not experienced during a training phase. 

To ameliorate these issues, methods have been developed to augment RL with uncertainty quantification (UQ). In this endeavor it has proven useful to distinguish two types of uncertainty \cite{Kiureghian2009,ABDAR2021243,DBLP:journals/ml/HullermeierW21}. \emph{Aleatoric} uncertainty stems from intrinsic randomness in the environments and cannot be reduced or eliminated by gathering more data through exploration. \emph{Epistemic} uncertainty, on the other hand, represents limitation of knowledge---the degree of lack of familiarity. It is akin to \emph{novelty}. Epistemic uncertainty \emph{can} be reduced through exploration.%
\footnote{Aleatoric (epistemic) uncertainty is also known as \emph{intrinsic (parametric)} uncertainty, respectively \cite{DBLP:journals/corr/abs-1710-10044,DBLP:conf/icml/MavrinYKWY19,DBLP:conf/iclr/NikolovKBK19,DBLP:journals/corr/abs-2109-03443}.} RL with UQ capabilities have demonstrated impressive empirical performance. An example is distributional RL \cite{DBLP:journals/corr/BellemareDM17,DBLP:journals/corr/abs-1710-10044,pmlr-v80-dabney18a}, which estimates not only the expected value but the probabilistic distribution of future returns. The identification of future risks that may result from current actions could be beneficial in high-stakes areas such as healthcare and autonomous driving. While incorporation of the aforementioned two kinds of uncertainty has been successful in RL with discrete actions \cite{DBLP:conf/icml/MavrinYKWY19,DBLP:journals/corr/abs-1905-09638,DBLP:conf/aaai/KeramatiDTB20}, the generalization to the realm of continuous actions has been only partially explorered \cite{DBLP:journals/corr/abs-1804-08617,DBLP:conf/nips/CiosekVLH19}. 

In this work, we propose a model-free off-policy RL algorithm for continuous control tasks that efficiently learns a risk-aware policy by taking both aleatoric and epistemic uncertainty into account. Our algorithm builds on the Deep Deterministic Policy Gradient algorithm (DDPG) \cite{DBLP:journals/corr/LillicrapHPHETS15}, one of the most popular RL methods for continuous actions. DDPG was selected as our basis due to its implementational simplicity. We extend DDPG by introducing multiple distributional critic networks. At the cost of increased gradient steps for training, our variant of DDPG enjoys quicker exploration of the environment, risk-aware policy learning, and enhanced trustworthiness through visualization of uncertainty to human eyes. Our approach contributes to solving industrial problems, where continuous actions are more the rule than the exception. 

In section~\ref{sc:lit} we review preceding works that address UQ for deep RL. In section~\ref{sc:pre} basic concepts of RL are summarized and the notation and terminology are fixed. In section~\ref{sc:method} our algorithm is formulated. In section~\ref{sc:bench} benchmark results are presented. We conclude in section~\ref{sc:end}.

\section{Background and related work\label{sc:lit}}
Treatment of uncertainty in RL is an extensively studied area of research. (A partial list of references related to our work can be found in Table~\ref{tb:ref}.) As early as in 1982, the variance and higher moments of the return distribution in generic Markov decision processes were studied \cite{sobel_1982}. Temporal-difference learning for the variance of return was elaborated in \cite{DBLP:conf/icml/TamarCM13,DBLP:journals/jmlr/TamarCM16}. In the context of RL, researches \cite{DBLP:conf/aaai/DeardenFR98,DBLP:conf/uai/MorimuraSKHT10,DBLP:conf/icml/MorimuraSKHT10} have attempted to evaluate the return distribution. More recently, the  distributional RL framework for deep RL was established by \cite{DBLP:journals/corr/BellemareDM17,DBLP:journals/corr/abs-1710-10044,DBLP:conf/icml/RowlandDKMBD19}. While QR-DQN \cite{DBLP:journals/corr/abs-1710-10044} assumed a predefined discrete set of \emph{quantile} points, it was soon extended to arbitrary quantiles \cite{pmlr-v80-dabney18a,DBLP:conf/nips/YangZLQ0L19}. Such distributional variants of DQN have shown strong performance in benchmark environments. The distributional setting was subsequently generalized to actor-critic models  \cite{DBLP:journals/corr/abs-1804-08617,Duan2021,2020arXiv200414547M,DBLP:conf/icml/KuznetsovSGV20,DBLP:conf/icml/NamKP21}, enabling applications to continuous-action problems. 

\begin{table*}[htb]
    \setlength\tabcolsep{-2pt}
    \caption{\label{tb:ref}An incomplete list of research papers on distributional RL and multi-actor/multi-critic RL. Acronyms that are used in different meanings are discriminated by prime $(')$.}
    \centering 
    \scalebox{0.9}{
    \begin{tabular}{|c|c||c|c|c|c|c|c|c|c|}
    	   \hline 
    	   \multicolumn{2}{|c||}{Feature} &
    	   \multicolumn{4}{c|}{
    	       Learn distributional value functions${}^{\mathstrut}$
    	   } & 
    	   \multicolumn{3}{c|}{
    	       Train multiple Q (critic) networks${}^{\mathstrut}$
    	   }
    	   & \multirow{2}{*}[-1.2em]{$\begin{array}{c}\text{Train multiple${}^{\mathstrut}$}
    	   \\\text{actor networks}\end{array}$}
    	   \\
    	   \cline{1-9}
        \multicolumn{2}{|c||}{Purpose} &
        $\begin{array}{c}
            \text{To boost}
            \\
            \text{performance}
            \\
            \text{(miscellaneous)}
            \end{array}$
        &
        $\begin{array}{c}
    	       \text{To suppress${}^{\mathstrut}$}
    	       \\
    	       \text{overestimation}\\\text{bias}
    	       \end{array}$
        &
        $\begin{array}{c}
    	       \text{To promote${}^{\mathstrut}$}
    	       \\
    	       \text{exploration}
    	       \end{array}$
        &
        $\begin{array}{c}
    	       \text{To learn a${}^{\mathstrut}$}
    	       \\
    	       \text{risk-aware policy}
    	       \end{array}$
        &
        $\begin{array}{c}
    	       \text{To suppress${}^{\mathstrut}$}
    	       \\
    	       \text{overestimation}\\\text{bias}
    	       \end{array}$
        &
        $\begin{array}{c}
    	       \text{To promote${}^{\mathstrut}$}
    	       \\
    	       \text{exploration}
    	       \end{array}$
    	   & 
    	   Other
    	   &
        \\\hline\hline 
        \multirow{2}{*}[-3.2em]{$\begin{array}{c}\text{Action}\\
        \text{space}\end{array}$} & Discrete${}^{\mathstrut}$ 
        &
        $\begin{array}{c}
            \text{C51 \cite{DBLP:journals/corr/BellemareDM17}}^{\mathstrut}\\
            \text{QR-DQN \cite{DBLP:journals/corr/abs-1710-10044}}\\
            \text{QR-A2C \cite{DBLP:journals/corr/abs-1806-06914}}\\
            \text{IQN \cite{pmlr-v80-dabney18a}}\\
            \text{ER-DQN \cite{DBLP:conf/icml/RowlandDKMBD19}}\\
            \text{MoG-DQN \cite{Choi2019ICRA}}\\
            \text{FQF \cite{DBLP:conf/nips/YangZLQ0L19}}\\
            \text{GMAC \cite{DBLP:conf/icml/NamKP21}}
        \end{array}$
        &
        ---
        & 
        $\begin{array}{c}
            \text{IDS \cite{DBLP:conf/iclr/NikolovKBK19}}\\
            \text{DLTV \cite{DBLP:conf/icml/MavrinYKWY19}}\\
            \text{CVaR-MDP \cite{DBLP:conf/aaai/KeramatiDTB20}}\\
            \text{NDQFN \cite{DBLP:conf/ijcai/ZhouZKZ21}}
        \end{array}$
        & 
        $\begin{array}{c}
            \text{UA-DQN \cite{DBLP:journals/corr/abs-1905-09638}}\\ 
            \text{CVaR-MDP \cite{DBLP:conf/aaai/KeramatiDTB20}}\\
            \text{SENTINEL \cite{DBLP:journals/corr/abs-2102-11075}}\\
            \text{IQN \cite{pmlr-v80-dabney18a}}
        \end{array}$
        & \text{DDQN \cite{DBLP:conf/aaai/HasseltGS16}}
        & 
        $\begin{array}{c}
            \text{Bootstrapped}\\
            \text{DQN \cite{DBLP:conf/nips/OsbandBPR16}}\\
            \text{Q-Ensembles \cite{2017arXiv170601502C}}\\
            \text{IDS \cite{DBLP:conf/iclr/NikolovKBK19}}\\
            \text{UA-DQN \cite{DBLP:journals/corr/abs-1905-09638}}\\
            \text{NDQFN \cite{DBLP:conf/ijcai/ZhouZKZ21}}
        \end{array}$
        & ---
        & ---
        \\
        \cline{2-10}
        & ~Continuous${}^{\mathstrut}~$ 
        &
        $\begin{array}{c}
            \text{D4PG \cite{DBLP:journals/corr/abs-1804-08617}}\\
            \text{SDPG \cite{DBLP:journals/corr/abs-2001-02652}}\\
            \text{DSAC$'$ \cite{2020arXiv200414547M}}\\
            \text{GMAC \cite{DBLP:conf/icml/NamKP21}}
        \end{array}$
        &
        $\begin{array}{c}
            \text{TQC \cite{DBLP:conf/icml/KuznetsovSGV20}} \\
            \text{DSAC \cite{Duan2021}} \\
            \text{ACC \cite{DBLP:journals/corr/abs-2111-12673}}
        \end{array}$
        & ---
        &
        $\begin{array}{c}
            \text{WCPG \cite{DBLP:conf/corl/TangZS19}} \\
            \text{DSAC$'$ \cite{2020arXiv200414547M}} \\
            \text{SDPG+CVaR \cite{DBLP:conf/l4dc/SinghZC20}} \\
            \text{O-RAAC \cite{DBLP:conf/iclr/UrpiC021}} \\
            \text{RC-DSAC \cite{Choi2021}} \\
            \text{\textbf{This work}}
        \end{array}^{\mathstrut}$
        & 
        $\begin{array}{c}
            \text{OAC \cite{DBLP:conf/nips/CiosekVLH19}} \\
            \text{TQC \cite{DBLP:conf/icml/KuznetsovSGV20}} \\
            \text{ADER \cite{DBLP:journals/corr/abs-2109-03443}} \\
            \text{ACC \cite{DBLP:journals/corr/abs-2111-12673}}
        \end{array}$
        & 
        $\begin{array}{c}
            \text{OAC \cite{DBLP:conf/nips/CiosekVLH19}} \\
            \text{ADER \cite{DBLP:journals/corr/abs-2109-03443}} \\
            \text{SOUP \cite{DBLP:conf/ijcai/ZhengYLCW18}} \\
            \text{SUNRISE \cite{DBLP:conf/icml/LeeLSA21}} \\
            \text{ED2 \cite{DBLP:journals/corr/abs-2111-15382}} \\
            \text{\textbf{This work}}
        \end{array}$
        &
        $\begin{array}{c}
            \text{MA-BDDPG} \\ \text{\cite{DBLP:conf/corl/KalweitB17}}\\
            \text{ACE \cite{DBLP:journals/corr/abs-1712-08987}}
            \end{array}$
        &
        $\begin{array}{c}
            \text{ACE \cite{DBLP:journals/corr/abs-1712-08987}}\\
            \text{ACE$'$ \cite{DBLP:conf/aaai/ZhangY19}}\\
            \text{SOUP \cite{DBLP:conf/ijcai/ZhengYLCW18}}\\
            \text{SUNRISE \cite{DBLP:conf/icml/LeeLSA21}}\\
            \text{ED2 \cite{DBLP:journals/corr/abs-2111-15382}}\\
            \text{\textbf{This work}}
            \end{array}$
        \\\hline 
    \end{tabular}
    }
\end{table*}

Distributional RL helps us address multiple fundamental goals in RL. First, it helps to ameliorate the so-called overestimation bias in Q learning \cite{DBLP:conf/nips/Hasselt10}. While it may be effectively resolved by Double DQN in the domain of discrete actions \cite{DBLP:conf/aaai/HasseltGS16}, the problem persists for continuous actions and people usually combat it with clipped double Q learning \cite{DBLP:conf/icml/FujimotoHM18}, which takes the minimum of two Q networks that are trained in parallel. Although this method yields significant gain in performance, it suffers from pessimistic underexploration. Recent works \cite{DBLP:conf/icml/KuznetsovSGV20,Duan2021,DBLP:journals/corr/abs-2111-12673} have pointed out that distributional value estimation could enable to elegantly overcome the overestimation bias problem without relying on clipped double Q learning. 

Secondly, distributional RL provides a very clear perspective on risk-sensitive decision making. Risk-sensitive RL (also known as risk-averse RL, safe RL, and conservative RL) is a field in which the RL agent is required to satisfy some safety constraints during training and/or deployment processes \cite{DBLP:journals/jmlr/GarciaF15}. Traditionally, variance of return has been conceived as the primary source of risk, and methods have been developed to obtain a policy that optimizes the average return corrected by variance \cite{DBLP:conf/sab/DilokthanakulS18,DBLP:conf/aaai/ZhangLW21}. Ref.~\cite{zbMATH01764831} proposed a risk-averse Q learning, the target of which actually coincides with the \emph{expectiles} of the return distribution \cite{DBLP:conf/icml/RowlandDKMBD19}. Distributional RL, which provides access to the entire return distribution, obviously gives much richer information than variance and makes it straightforward to optimize policies with respect to various risk-sensitive criteria such as CPW \cite{Tversky1992}, Sharpe ratio \cite{10.2307/2351741}, Value at Risk (VaR) \cite{Jorion2006}, and Conditional Value at Risk (CVaR) \cite{Rockafellar2000,10.5555/2969033.2969218}. See  \cite{DBLP:conf/icml/MavrinYKWY19,DBLP:journals/corr/abs-1905-09638,DBLP:conf/aaai/KeramatiDTB20,DBLP:journals/corr/abs-2102-11075,pmlr-v80-dabney18a} for works on risk-sensitive distributional RL with discrete actions. Generalization to the domain of continuous actions has subsequently followed \cite{2020arXiv200414547M,DBLP:conf/l4dc/SinghZC20,DBLP:conf/iclr/UrpiC021,Choi2021}. 

Thirdly, distributional RL facilitates exploration. It is well known that efficient exploration in high-dimensional state/action space is quite challenging. It is even harder in sparse-reward environments. Standard methods such as the \text{$\varepsilon$-greedy} policy, addition of noise to the action, and entropy regularization of the policy, often turn out to be insufficient and miscellaneous alternative approaches have been developed, as reiewed in \cite{DBLP:journals/corr/abs-2109-00157,DBLP:journals/corr/abs-2109-06668}. One of effective ways to encourage exploration is to take epistemic uncertainty into an RL algorithm explicitly. Although the breadth of the return distribution after training mainly originates from aleatoric uncertainty, in early stages of training the return distribution can comprise \emph{both} aloatoric and epistemic uncertainty. As a result, the addition of a variance-related piece to the reward function induces efficient exploration \cite{DBLP:conf/sab/DilokthanakulS18,DBLP:conf/icml/MavrinYKWY19}. In the same vein, artificially shifting the return distribution to a higher value forces the agent to thoroughly explore the environment optimistically \cite{DBLP:conf/aaai/KeramatiDTB20}. Although their technical simplicity is appealing, these methods suffer from the downside that the two types of uncertaity remain entangled during training and cannot be controlled independently. To fix this problem, Refs.~\cite{DBLP:conf/iclr/NikolovKBK19,DBLP:journals/corr/abs-1905-09638} have proposed to concurrently train multiple Q networks to extract epistemic uncertainty from the discrepancy of Q values, while making each Q network distributional so that aleatoric uncertainty can be estimated separately. This approach allows us to accelerate training for learning a risk-aware policy, or put it differently, ``Behave optimistically during training and act conservatively after deployment.'' However, the aforementioned works \cite{DBLP:conf/iclr/NikolovKBK19,DBLP:journals/corr/abs-1905-09638,DBLP:conf/aaai/KeramatiDTB20} are all restricted to environments with discrete actions.

Training an ensemble of deep neural networks is a popular approach to epistemic uncertainty quantification and out-of-distribution detection \cite{NIPS2017_9ef2ed4b,2021arXiv210914117L}. It can be viewed as performing an approximate Bayesian inference \cite{Pearce1805,Pearceaistats2020,2021arXiv210513283H}. Its effectiveness for hard exploration problems in deep RL has been demonstrated in \cite{DBLP:conf/nips/OsbandBPR16}. While the diversity of members is a key to the success of ensembling in machine learning \cite{2021arXiv210914117L}, empirical studies \cite{DBLP:conf/nips/OsbandBPR16,NIPS2017_9ef2ed4b} verified that training neural networks with independent random initializations on the same dataset without bootstrapping could yield sufficiently good performance. We remark that, although state-of-the-art RL algorithms for continuous control such as TD3 \cite{DBLP:conf/icml/FujimotoHM18} and SAC \cite{DBLP:conf/icml/HaarnojaZAL18} train two critic networks concurrently, they do so to reduce the overestimation bias of Q learning rather than to promote exploration. Papers that studied actor-critic ensemble approaches to continuous control problems are listed in the bottom-right cells of Table~\ref{tb:ref}.

In this work, motivated by these lines of work, we propose a model-free off-policy actor-critic algorithm that allows us to simultaneously deal with both types of uncertainties in continuous action spaces. Specifically, we present an extension of distributional DDPG \cite{DBLP:journals/corr/LillicrapHPHETS15,DBLP:journals/corr/abs-1804-08617} to the architecture with multiple actors and critics, thereby encouraging the agent to take exploratory actions based on disagreement of critics, while learning a conservative policy based on distributional estimates of returns, at the same time. Our method, named \emph{Uncertainty-Aware DDPG} (UA-DDPG), aims to not only reduce sample complexity of training but also bring more flexibility to policy optimization criteria. 

Note that UA-DDPG is different from other actor-critic algorithms built upon critic ensembles \cite{DBLP:conf/corl/KalweitB17,DBLP:journals/corr/abs-1712-08987,DBLP:conf/ijcai/ZhengYLCW18,DBLP:conf/nips/CiosekVLH19,DBLP:journals/corr/abs-2109-03443,DBLP:conf/icml/LeeLSA21,DBLP:journals/corr/abs-2111-15382} which completely ignore aleatoric uncertainty. Furthermore, in contrast to most of multi-critic approaches that enhance exploration either by way of Thompson sampling \cite{DBLP:conf/nips/OsbandBPR16,DBLP:journals/corr/abs-1905-09638,DBLP:conf/ijcai/ZhengYLCW18,DBLP:journals/corr/abs-2111-15382} or by adding an exploration bonus to the reward or Q values \cite{2017arXiv170601502C,DBLP:conf/icml/LeeLSA21,DBLP:conf/ijcai/ZhouZKZ21,DBLP:journals/corr/abs-2109-03443}, UA-DDPG selects actions that directly  maximize epistemic uncertainty. It bears similarity to OAC \cite{DBLP:conf/nips/CiosekVLH19}, which slightly modifies the action according to optimistically calculated Q values. However, we do not impose constraints on the distance between the target policy and the behavior policy as in \cite{DBLP:conf/nips/CiosekVLH19}. It is also noteworthy that some of the preceding works \cite{DBLP:journals/corr/abs-1905-09638,DBLP:conf/ijcai/ZhouZKZ21,DBLP:conf/nips/CiosekVLH19,DBLP:journals/corr/abs-2109-03443} are formulated for the case of two critics only, whereas UA-DDPG works with a general $(\geq 2)$ number of critics. Finally, we remark that although TQC \cite{DBLP:conf/icml/KuznetsovSGV20} and ACC \cite{DBLP:journals/corr/abs-2111-12673} combine distributional RL with ensemble learning like UA-DDPG, they lack a mechanism to expedite exploration by using epistemic uncertainty---they only use the ensemble of critics to suppress the overestimation bias of Q learning.

\section{Preliminaries\label{sc:pre}}

As an environment for RL, we consider a Markov Decision Process (MDP) defined by a tuple $(\mathcal{S},\mathcal{A},r, P,\gamma,p_0)$. $\mathcal{S}$ is the state space, $\mathcal{A}$ is the action space, $r:\mathcal{S}\times\mathcal{A}\times\mathcal{S}\to \mathbb{R}$ is the reward function $r(s_t, a_t, s_{t+1})\in\mathbb{R}$, $P:\mathcal{S}\times\mathcal{A}\times\mathcal{S}\to\mathbb{R}$ is the transition probability $P(s_{t+1}|s_t, a_t)\geq 0$, $\gamma\in[0,1]$ is the discount factor, and $p_0:\mathcal{S}\to\mathbb{R}$ is the initial state distribution $p_0(s)\geq 0$. Let $\pi:\mathcal{S}\times\mathcal{A}\to \mathbb{R}$ denote a policy of an RL agent with the action probability density $\pi(a|s) \geq 0$. The target of optimization in RL is the expectation value of the discounted long-term cumulative rewards, viz.~return, given by
\ba
    \mathbb{E}_{\pi,P}\kkakko{\sum_{t=0}^{\infty}\gamma^t r(s_t,a_t,s_{t+1})}
    \label{eq:rrr}
\ea
where $a_t\sim \pi(\cdot|s_t)$, $s_{t+1}\sim P(\cdot|s_t, a_t)$ and $s_0\sim p_0$. The task is to find an optimal policy $\pi^*$ that maximizes \eqref{eq:rrr}. 

The Q function $Q^\pi:\mathcal{S}\times\mathcal{A}\to\mathbb{R}$ describes the gain of return by taking a certain action in a certain state and afterwards following the policy $\pi$,
\ba
	Q^\pi(s,a)& = \mathbb{E}_{\pi,P}\kkakko{
	    \sum_{t=0}^{\infty}\gamma^t r(s_t, a_t, a_{t+1}) \Big| s_0=s, a_0=a
	}.\hspace{-2mm}
\ea
It satisfies the Bellman equation \cite{doi:10.1126/science.153.3731.34}
\ba
    Q^\pi(s,a) = \mathbb{E}_P \kkakko{r(s,a,s')} 
    + \gamma \mathbb{E}_{\pi,P} \kkakko{Q^\pi(s',a')}
\ea
where $s'\sim P(\cdot | s,a)$ and $a'\sim \pi(\cdot|s')$.

In the distributional RL setup \cite{DBLP:journals/corr/BellemareDM17,DBLP:journals/corr/abs-1710-10044,pmlr-v80-dabney18a,DBLP:conf/icml/RowlandDKMBD19} we consider a random variable version of the Q function $Z^\pi(s,a)$, such that $Q^\pi(s,a)=\mathbb{E}_{\pi,P}[Z^\pi(s,a)]$. The distributional Bellman equation reads
\ba
    Z^\pi(s,a) \overset{D}{=} r(s,a,s') + \gamma Z^\pi(s', a')\,,
\ea
where $\overset{D}{=}$ implies that both sides of the equation obey the same law of probability, with the understanding that $s'\sim P(\cdot|s,a)$ and $a'\sim\pi(\cdot | s')$. From here the Bellman optimality equation readily follows if $a'$ is chosen such that
\ba
	a'=\underset{\wt{a}\in\mathcal{A}}{\mathrm{arg\,max}} ~
	\mathbb{E}_{\pi,P}[Z^\pi(s',\wt{a})]\,.
\ea
We follow \cite{DBLP:journals/corr/abs-1710-10044} and define $N$ quantile points over $[0,1]$ as
\ba
	\wh{\tau}_i = \frac{2i-1}{2N}, \quad i=1,2,\dots,N.
\ea
Let $z_{\tau}(s,a)$ denote the $\tau$-quantile of the cumulative probability distribution of $Z^\pi(s,a)$. If a neural network is used to model $\{z_\tau\}:\mathcal{S}\times\mathcal{A}\times[0,1]\to\mathbb{R}$, the network parameters shall be updated by taking gradient descent of the critic loss function
\ba
	\mathcal{L}_C & = \frac{1}{|B|}\sum_{(s,a,r,s')\in B}
	\frac{1}{N^2}\sum_{i=1}^{N}\sum_{j=1}^{N}
	\rho^{\kappa}_{\wh{\tau}_i}(\Delta_{ij})\,,
	\label{eq:crloss}
	\\
	\Delta_{ij} & = r + \gamma z_{\wh{\tau}_j}(s',a') 
	- z_{\wh{\tau}_i}(s,a)\,, \quad a'\sim \pi(\cdot|s') 
	\label{eq:tderr}
\ea
where $B$ is a minibatch of transitions sampled from a replay buffer and $\rho$ is a quantile Huber loss given by
\ba
	\rho_\tau^\kappa(x) & = |\tau-\delta_{\{x<0\}}| \mathcal{L}_\kappa(x)\,,
	\\
	\mathcal{L}_\kappa(x) & =\begin{cases}\frac{1}{2}x^2, 
	\qquad \qquad ~\text{if $|x|\leq \kappa$} \\ 
	\kappa\mkakko{|x|-\frac{1}{2}\kappa} \quad \text{otherwise}\end{cases}
	\hspace{-3mm},
\ea
where $\delta_{\{\clubsuit\}}=1$ if $\clubsuit$ is true and 0 otherwise. 
When $\mathcal{A}$ is a finite discrete set, the optimal action to be substituted into $a'$ in \eqref{eq:tderr} is simply given by the maximizer of the average return
\ba
	a' = \underset{\wt{a}\in\mathcal{A}}{\mathrm{arg\,max}}\;
	\frac{1}{N}\sum_{i=1}^{N} z_{\wh{\tau}_i}(s',\wt{a})\,.
\ea
When $\mathcal{A}$ is a continuous domain, and the policy is deterministic,%
\footnote{We switch from $\pi$ to $\mu$ to conform to the convention in the literature.} the parameters of the policy network $\mu:\mathcal{S}\to\mathcal{A}$ may be updated via gradient descent of the loss function \cite{DBLP:journals/corr/abs-1804-08617}
\ba
	\mathcal{L}_A = - \frac{1}{|B|}\sum_{s \in B} \frac{1}{N}
	\sum_{i=1}^{N}z_{\wh{\tau}_i} \mkakko{s,\mu(s)}\,.
	\label{eq:LAneut}
\ea
If the policy is stochastic, actions should be sampled using the reparametrization trick \cite{DBLP:journals/corr/KingmaW13}.

Note that the loss function \eqref{eq:LAneut} is tailored for training of a \emph{risk-neutral} policy. We can write it as $\mathcal{L}_A \propto - \underset{\tau\sim U([0,1])}{\mathbb{E}}[z_\tau]$, 
where $U([0,1])$ is a uniform probability distribution on $[0,1]$. In the case of a risk-aware policy, we need to modify it as \cite{pmlr-v80-dabney18a,2020arXiv200414547M} $\mathcal{L}_A \propto - \underset{\tau\sim U([0,1])}{\mathbb{E}}\big[z_{\beta(\tau)}\big]$ where $\beta:[0,1]\to[0,1]$ is a function describing the sensitivity to risk. For instance, the CVaR$_\eta$ criterion with $0<\eta<1$ corresponds to $\beta(\tau)=\eta \tau$. A small value of $\eta$ would indicate strong risk aversion and the limit $\eta\to +0$ is akin to the worst case analysis in robust optimization \cite{GORISSEN2015124}. Although the choice of $\beta$ does not modify the critic loss \eqref{eq:crloss} explicitly, the action $a'$ in the temporal-difference error \eqref{eq:tderr} must be selected according to the $\beta$-dependent policy.

\section{Method\label{sc:method}}

In the following, we introduce the \emph{Uncertainty-Aware Deep Deterministic Policy Gradient} algorithm (UA-DDPG), designed for continuous control with both aleatoric and epistemic uncertainties taken into account for flexible risk modeling, faster exploration and out-of-distribution monitoring.
\begin{wrapfigure}{r}{.57\columnwidth}
    \centering
    \vspace{-5pt}
    \includegraphics[width=.57\columnwidth]{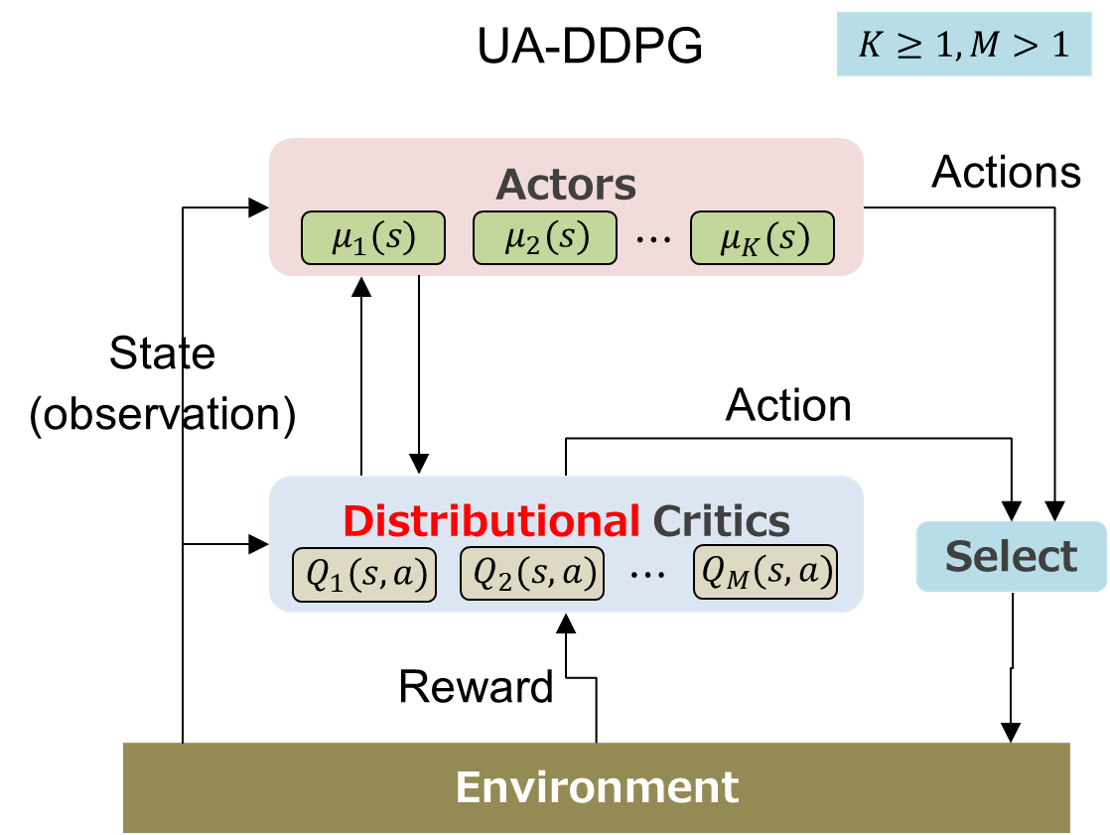}
    \vspace{-13pt}
    \caption{\label{fg:34543}Architecture of the UA-DDPG agent.}
    \vspace{-7pt}
\end{wrapfigure}
The architecture is shown in Figure~\ref{fg:34543}. In addition to the $K$ actors and $M$ critics shown in the figure, there are $K+M$ \emph{target networks} (not shown) for both actors and critics, which slowly keep track of changes in the trained networks via Polyak averaging. 
Each actor $\mu:\mathcal{S}\to\mathcal{A}$ is denoted as $\mu(\theta_k)$, where $\theta_k$ is the set of parameters of the $k$th actor. Each critic has $N$ quantile outputs $Q_i(\phi_m)$, $i=1,\cdots,N$, where $\phi_m$ is the set of parameters of the $m$th critic. The parameters of the target networks will be denoted as $\wh{\theta}_k$ and $\wh{\phi}_m$. At the beginning of training, parameters of all trained neural networks are independently and randomly initialized, with values drawn from the Gaussian distribution $\mathcal{N}(0,\sigma^2)$. The variance $\sigma^2$ is an important hyperparameter of UA-DDPG. We have also tested other initialization schemes \cite{10.5555/3327757.3327952,Pearceaistats2020} but did not observe improvement in performance.

UA-DDPG is an off-policy algorithm. All interactions with the environment are saved in a replay buffer and old data are constantly deleted. Actors and critics are trained on minibatches of transitions drawn from the replay buffer. 
\paragraph{Actor loss} 
Actors are trained to maximize a risk-sensitive average of the quantiles of critics. The latter may be written, without loss of generality, as $\sum_{i=1}^{N}\beta_i Q_i$. In the risk-neutral case, we have $\beta_1=\beta_2=\cdots=\beta_N=1/N$; for CVaR$_\eta$ we have $\beta_i=\delta_{\{\!\frac{i}{N}\leq \eta\}}/(N\eta)$. The loss for the $k$th actor can now be explicitly stated as
\ba
	\mathcal{L}_A(\theta_k) & = - \frac{1}{|B|}\sum_{s\in B}
	\sum_{i=1}^{N}\beta_i \overline{Q}_i^{(k)}(s) \,,
	\label{eq:LAdsfdf4}
	\\
	\overline{Q}^{(k)}_i(s) & = \frac{1}{M}\sum_{m=1}^{M} 
	Q_i(s, a | \phi_m)\Big|_{a=\mu(s|\theta_k)} \,. 
	\label{eq:Qfg}
\ea
In taking the gradient steps of the above loss, the critic parameters $\{\phi_m\}_{m=1}^{M}$ should be held constant. 
\paragraph{Critic loss}
The loss function for the $m$th critic is given by an adaptation of \eqref{eq:crloss} to the case of multi-critics as
\ba
	\mathcal{L}_C(\phi_m) & = \frac{1}{|B|}\sum_{(s,a,r,s')\in B}
	\frac{1}{N^2}\sum_{i=1}^{M}\sum_{j=1}^{M}
	\rho^{\kappa}_{\wh{\tau}_i}(\Delta_{ij})\,,
	\label{eq:LC08789d7sfsd}
	\\
	\Delta_{ij} & = r + \gamma \overline{\overline{Q}}_j(s') - Q_i(s,a| \phi_m)\,,
	\\
	\overline{\overline{Q}}_j(s') & = \frac{1}{M}\sum_{n=1}^{M}
	Q_j(s',a'|\wh{\phi}_n)\Big|_{a'=\mu(s'|\wh{\theta}_{k^*})}\,.
	\label{eq:2345435}
\ea
The actor index $k^*$ in the last line is defined as 
\ba
	k^* & = \underset{1\leq k \leq K}{\mathrm{arg\,max}}
	\sum_{i=1}^{N}\beta_i \kkakko{\overline{Q}_i^{(k)}(s')}^t \,,
\ea
where the superscript $t$ indicates that it is the ``target network version,'' namely, 
$\kkakko{\overline{Q}_i^{(k)}(s')}^t$ is same as $\overline{Q}^{(k)}_i(s')$ 
but with $\phi$ and $\theta$ in \eqref{eq:Qfg} replaced with $\wh{\phi}$ and 
$\wh{\theta}$, respectively. 

Note that $\overline{\overline{Q}}_j(s')$ in \eqref{eq:2345435} does not depend on $m$;  the Bellman target is the same for all critics. Thus, as the training proceeds, the predictions of all critics will converge for such state-action pairs that are stored in the buffer and have been sampled sufficiently many times. Then the disagreement across critics would serve as an indicator of epistemic uncertainty, or simply \emph{novelty}, of a given state-action pair. 

It is worthwhile to emphasize that a situation with high epistemic uncertainty carries at least two implications. First, during training, it is beneficial for the agent to experience such a situation to expand knowledge and improve the policy, so the agent is encouraged to seek high epistemic uncertainty. Second, after deployment in a real environment, encounter with a situation with unexpectedly high epistemic uncertainty should raise a warning, since it is out-of-distribution---the agent has not experienced it before and its next action is likely to be suboptimal or even harmful. The human user of AI may find it necessary to collect data and send the agent to a new training phase in order to keep up with novel changes in the environment. 

With these remarks in mind, we now proceed to the action selection scheme during training. In off-policy learning, it is possible in principle to collect data with a behavior policy that is quite dissimilar to the target policy. However, with such excessive off-policiness the agent runs the risk of learning a suboptimal policy and/or prolonged training time. A quick fix to this in the $\varepsilon$-greedy method is to let $\varepsilon$ decay over time. Motivated by this, we define a scheduling function 
\ba
	p(t) & = \max\mkakko{1 - \frac{t}{T_{\exp}},~p_{\min}} 
\ea
where $t\geq 0$ is the time steps in the environment, $T_{\exp}$ is the time scale for exploration, and $p_{\min}\in[0,1]$ is the minimal exploration rate at late times. In the training phase of UA-DDPG, we let the agent take a greedy action with probability $1-p(t)$ and an exploratory action with probability $p(t)$. Now the greedy action in state $s$ is given by $\mu(s|\theta_{k^{**}})$, where the ``best'' actor index $k^{**}$ is determined from
\ba
	k^{**} & = \underset{1\leq k \leq K}{\mathrm{arg\,max}}
	\sum_{i=1}^{N}\beta_i \overline{Q}_i^{(k)}(s) 
	\label{eq:fgf6545465}
\ea
with $\overline{Q}_i^{(k)}(s)$ defined in \eqref{eq:Qfg}. To promote exploration we will add a small Gaussian noise to the greedy action as suggested in \cite{DBLP:journals/corr/LillicrapHPHETS15}. It is not \emph{a priori} clear whether the above actor selection scheme is best or not. We have therefore tested options (i) and (ii), where (i) is to select an actor randomly at every time step, and (ii) is to select an actor randomly at the start of every episode and continue using that actor till the end of the episode. However, we observed only mediocre performance for these options in numerical experiments and hence stopped pursuing them further. Our greedy scheme \eqref{eq:fgf6545465} is in line with \cite{DBLP:journals/corr/abs-1712-08987,DBLP:conf/aaai/ZhangY19,DBLP:conf/icml/LeeLSA21} but at odds with \cite{DBLP:conf/ijcai/ZhengYLCW18,DBLP:journals/corr/abs-2111-15382}. 

Using multiple actors is expected to boost performance, for a Q function in action space is likely to be highly multimodal and it would not be easy to escape from a local optimum if only a single actor is used. That said, higher $K$ will inevitably increase the computational cost and, if this is problematic, nothing prevents us from setting $K=1$. UA-DDPG is well-defined for both $K=1$ and $K>1$.

Next we move on to define an exploratory action, which is taken with probability $p(t)$ during training. Let us employ the variance of multiple critics for a given state-action pair as approximate Epistemic Uncertainty:
\ba
    \mathrm{EU}(s,a) & = \frac{1}{N}\sum_{i=1}^{N}
    \mathrm{Var}\Big[ \big\{Q_i(s,a|\phi_m) \big\}_{1\leq m \leq M}\Big]\,.
\ea
Suppose the agent is in state $s$ now. The action that is \emph{maximally exploratory} would be $\underset{a\in\mathcal{A}}{\mathrm{arg\,max}}\;\mathrm{EU}(s,a)$, but solving this nonlinear continuous optimization problem at every time step is too time-consuming. As a cheap alternative, we consider a one-dimensional space $\Omega(s)\subset\mathcal{A}$ defined as
\ba
	\Omega(s) & = \ckakko{a \in\mathcal{A} \;| \;  
	a = \mu(s|\theta_{k^{**}}) + c \mathbf{V},~0\leq c <\infty}
	\label{eq:Ome}
\ea
where $\mu(s|\theta_{k^{**}})$ is the greedy action in state $s$, as noted before, and
\ba
	\mathbf{V} & = \nabla_{a}\mathrm{EU}(s,a)\Big|
	_{a=\mu(s|\theta_{k^{**}})}
	\label{eq:sdgfd8g989}
\ea
is a vector in the direction of steepest ascent of EU. The computational overhead associated with the gradient calculation using automatic differentiation is virtually negligible. In many cases the action range is finite and hence there is a natural upper limit on $c$ in \eqref{eq:Ome}. For clarification, we note that OAC \cite{DBLP:conf/nips/CiosekVLH19} used the gradient vector of an optimistic Q function to facilitate exploration, while our $\mathbf{V}$ is the gradient of the approximate epistemic uncertainty. 

While solving the original problem of finding $\underset{a\in\mathcal{A}}{\mathrm{arg\,max}}\;\mathrm{EU}(s,a)$ is not feasible, it is numerically inexpensive to search for the approximate maximum of $\mathrm{EU}(s,a)$ over $\Omega(s)$ by just computing $\mathrm{EU}(s,a)$ for finitely many $a$'s that are equidistributed over $\Omega(s)$. In this fashion, exploration of the environment is efficiently promoted. We do not add a noise to the exploratory action. Although it is conceivable that using more critics will benefit accurate estimation of epistemic uncertainty, it entails  significantly higher numerical costs proportional to $M$. An actual number of $M$ should be decided on the basis of available computational resources. 

In the inference time, the agent no longer takes exploratory actions---it only takes greedy actions $\mu(s|\theta_{k^{**}})$ without noise. However, computing aleatoric and epistemic uncertainties from the outputs of critics and communicating those values to humans could be useful from the viewpoint of quality assurance and accident prevention. 

The whole flow of UA-DDPG is summarized as a pseudo-code in Algorithm~\ref{alg:main}. A few important parameters are defined there as well.

\begin{algorithm}[t]
    \caption{Uncertainty-Aware Deep Deterministic Policy Gradient (UA-DDPG)}
    \label{alg:main}
    \begin{algorithmic}[1]
        \REQUIRE 
        $\{\{Q_i(\phi_m)\}_{i=1}^{N}\}_{m=1}^{M}$: $M\,(\geq 2)$ critic networks with $N$ quantile outputs, $\{\mu(\theta_k)\}_{k=1}^{K}$: $K\,(\geq 1)$ actor networks, $\{\beta_i\}_{i=1}^{N}$: coefficients that control risk sensitivity, $\sigma^2$: variance of initial network parameters, $\gamma\in[0,1]$: discount rate, $T_{\exp}$: duration of the exploratory phase, $p_{\min}\in[0,1]$: minimum exploration rate, $\sigma^2_{\mathcal{A}}$: variance of the Gaussian noise added to greedy actions, $S_0$: initial steps for random actions, $\eta_A$: learning rate for actors, $\eta_C$: learning rate for critics, $D$: replay buffer, $U_{\max}$: threshold of epistemic uncertainty
        \STATE Initialize actor and critic networks with parameters drawn from $\mathcal{N}(0,\sigma^2)$
        \STATE Prepare target networks with parameters $\wh{\theta}$ and $\wh{\phi}$ 
        copied from $\theta$ and $\phi$
        \STATE $D\leftarrow \emptyset$
        \FOR{$S_0$ environment steps}
        \STATE
        Take a random action, receive reward, and observe the next state 
        \STATE
        $D \leftarrow D \,\cup\,\{(s,a,r,s')\}$
        \ENDFOR
        \STATE Use $D$ to fix the normalization of rewards and states
        \FOR{each training iteration}
        \FOR{each environment step}
        \STATE 
        Compute $a_g=\mu(s|\theta_{k^{**}})$
        \hfill $\triangleright$ \eqref{eq:fgf6545465}%
        \STATE Sample $\alpha\sim U([0,1])$
        \IF{$\alpha > p(t)$}
        \STATE
        $a_g \leftarrow a_g + \varepsilon$, \quad 
        $\varepsilon\sim\mathcal{N}(0, \sigma^2_{\mathcal{A}})$
        \STATE
        Clip $a_g$ to the allowed range
        \STATE 
        Take action $a_g$ in the environment
        \ELSE
        \STATE Compute the gradient 
        $\mathbf{V}$ at $a=a_g$ \hfill $\triangleright$ \eqref{eq:sdgfd8g989}%
        \STATE 
        Approximately compute $a_e=\underset{a\in\Omega(s)}{\mathrm{arg\,max}}\;\mathrm{EU}(s,a)$
        \STATE Take action $a_e$ in the environment
        \ENDIF
        \STATE Receive reward and observe the next state 
        \STATE $D\leftarrow D\,\cup \, \{(s,a,r,s')\}$
        \ENDFOR
        \IF{time to update}
        \FOR{each gradient step}
        \STATE Sample a minibatch $B$ of trainsitions from $D$
        \STATE Compute loss $\mathcal{L}_A$ and $\mathcal{L}_C$
        \hfill $\triangleright$ \eqref{eq:LAdsfdf4} \& \eqref{eq:LC08789d7sfsd}%
        \STATE $\theta \leftarrow \theta - \eta_A \nabla_{\theta} \mathcal{L}_A$
        \STATE $\phi\leftarrow \phi - \eta_C \nabla_{\phi}\mathcal{L}_C$
        \ENDFOR 
        \STATE $\wh{\theta}\leftarrow \tau \wh{\theta}+(1-\tau)\theta$
        \hfill $\triangleright$ Update target networks
        \STATE $\wh{\phi}\leftarrow \tau \wh{\phi}+(1-\tau)\phi$
        \ENDIF
        \ENDFOR 
        \FOR{each inference step}
        \STATE Compute $a_g=\mu(s|\theta_{k^{**}})$
        \IF{$\mathrm{EU}(s,a_g) > U_{\max}$}
        \STATE Raise a warning
        \ENDIF 
        \STATE Take action $a_g$ in the environment, receive reward and observe the next state
        \ENDFOR
    \end{algorithmic}
\end{algorithm}

\section{Experimental results\label{sc:bench}}

To test the effectiveness of our algorithm, we have conducted numerical experiments in several simulation environments. The hyperparameters used for each simulation are summarized in the \hyperref[ap:4]{Appendix}. Unless stated otherwise, we have used the same hyperparameters for both UA-DDPG and baselines to assure a fair comparison. 

\begin{wrapfigure}{r}{100pt}
    \centering
    \includegraphics[width=90pt]{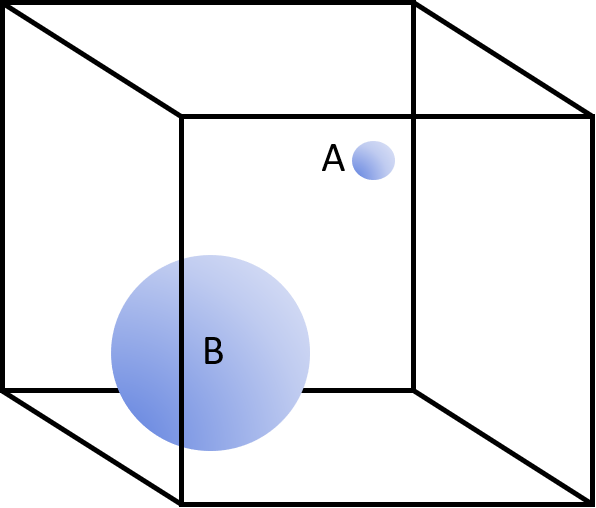}
    \vspace{-4pt}
    \caption{\label{fg:cube}A simple RL environment in $\mathbb{R}^3$.}
    \vspace{-14pt}
\end{wrapfigure}

\subsection{Exploration in a cube}

we consider a hard exploration problem as shown in Figure~\ref{fg:cube}. The state space of the agent is a cube $[-1,1]^3$ with periodic boundary conditions in all directions. The initial position of the agent is randomly chosen in the vicinity of the corner $(-1,-1,-1)$. At each timestep, the agent selects a velocity vector $v\in [-0.05,0.05]^3$ and moves forward in the direction of $v$. There is a ball “B” with a center $(-0.5,-0.5,-0.5)$ and radius $0.2$. The reward is equal to $-0.1$ if the agent is inside B, otherwise $-0.2$. The maximum episode length is 200. An episode is terminated immediately if the agent reaches the ball “A”, which is centered at $(0.3,0.2,0.1)$ with radius $0.06$. There is no randomness in this environment. 
\vspace{-2pt}

The agent always receives a negative reward; hence it is clear that the optimal policy is to go straight to the ball A from the starting position and terminate the episode as quickly as possible. However, this is a very tough task for the agent which has no clue about the location of the ball A; notice that the volume of the ball A is only $0.01\%$ of the cube. To find the ball A the agent must explore the interior of the cube thoroughly. The ball B is an attractive, albeit a suboptimal, goal for the agent as it gives a higher reward than the outside. 

In numerical simulations we noticed that this environment is too challenging for both DDPG and UA-DDPG. To mitigate the difficulty we adopted two tricks. The first is prioritized experience replay \cite{DBLP:journals/corr/SchaulQAS15}. The second is a periodical suspension of exploration. In UA-DDPG, exploratory actions that are completely decorrelated with the target policy are taken frequently, especially in the early stages of the training, which can hurt performance. To mitigate this ``excessive off-policiness,'' we turn off the action noise and suspend exploratory actions during every $S$-th episode, which we expect will weaken the bias of state distributions in the replay buffer. We stress that these tricks were used only for this environment and not used for others in the following subsections. 

\begin{figure}[htb]
	\centering
	\includegraphics[width=\columnwidth]{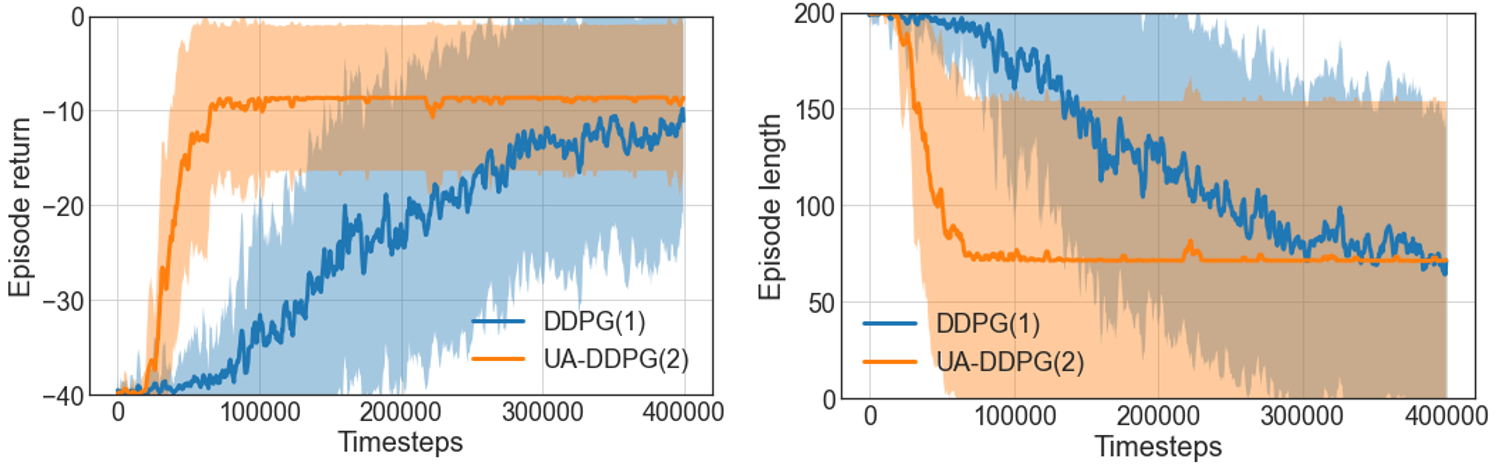}
	\vspace{-14pt}
	\caption{\label{fg:cubegraph}Results for the cubic environment. Solid lines are mean and the shaded areas are $\pm$ one standard deviation.}
\end{figure}

In numerical experiments we ran training with 24 random seeds to obtain statistically meaningful results. As there is no intrinsic randomness in this environment, we set $N=1$. The results are summarized in Table~\ref{tb:cube}. In addition, the learning curves of DDPG (No.1) and UA-DDPG (No.2) are shown in Figure~\ref{fg:cubegraph}.
\begin{table}[h]
	\caption{\label{tb:cube}Results for the cubic test environment. The best score is in bold. The errors represent one standard deviation.}
	\centering
	\begin{tabular}{cccccc}
		\hline 
		No. & Algorithm & $M$ & $K$ & $S$ & Return
		\\\hline
		1 & DDPG &1&1&---&$-12.4\pm 9.1$
		\\
		2 & UA-DDPG &3&4&8& {$\bf -8.8\pm 7.8$}
		\\
		3 & UA-DDPG &3&4&---&$-12.2\pm 8.4$
		\\
		4 & UA-DDPG &3&1&8&$-13.8\pm 10.1$
		\\
		5 & UA-DDPG &10&4&8&$-15.1\pm 8.1$
		\\
		6 & UA-DDPG &2&4&8&$-12.7\pm 8.6$
		\\\hline 
	\end{tabular}
\end{table}
Clearly UA-DDPG (No.~2) outperforms DDPG by a large margin, although variance is large for both algorithms. By the end of the first 70000 steps, about 71\% of UA-DDPG agents learned a nearly optimal policy, whereas DDPG agents needed 400000 steps to achieve the same level of performance. It amounts to 82\% reduction of the training time. 

A few additional remarks are in order.
\begin{itemize}
	\item
	Comparison of No.2 ($M=3$), No.5 ($M=10$) and No.6 ($M=2$) reveals that $M=3$ performs best, while $M=10$ performs worst. $M=2$ is in between. This non-monotonic dependence on $M$ is not intuitive for us and is left for future research.
	\item
	There is a large gap in scores between No.2 ($K=4$) and No.4 ($K=1$), showing that the use of multiple actors significantly boosts performance.
	\item
	The fact that No.2 ($S=8$) substantially outperforms No.3 ($S$ not used) highlights that periodical insertion of on-policy rollouts is indeed effective for mitigation of the off-policy bias.
\end{itemize}

\subsection{Robot locomotion task}

Robotics is one of the oldest areas of RL applications. RL offers a powerful framework to generate sophisticated behaviors of robots that are hard to engineer manually. In order to test the applicability of our approach in this domain, we have trained and tested a UA-DDPG agent using an OSS robotic simulator PyBullet%
\footnote{\url{https://pybullet.org/wordpress/}~\url{https://github.com/bulletphysics/bullet3}}. Among the locomotion tasks implemented in PyBullet, we chose 'HopperBulletEnv-v0', which is a single-foot robot with three joints. The dimension of the action space is 3 and the dimension of the state space is 15, including the $(x,y)$-position on the plane, the height of the body from the plane, angles of joints, and the velocity of joints. The reward is a sum of four terms: progress bonus, alive bonus, electricity cost, and cost of joints at limit. The reward gets higher if the agent learns to move forward faster. 

\begin{figure}[htb]
	\centering
	\includegraphics[width=.7\columnwidth]{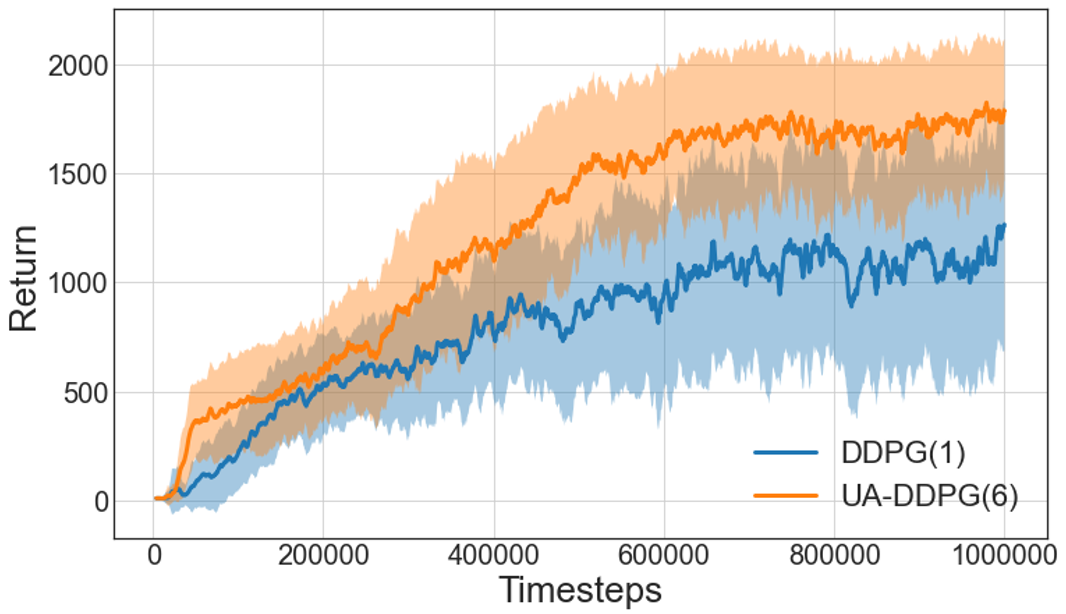}
	\vspace{-4pt}
	\caption{\label{fg:hopgraph}Learning curves in the PyBullet Hopper-v0 environment. Solid lines are mean and the shaded areas are $\pm$ one standard deviation.}
	\vspace{2mm}
	\centering
	\includegraphics[width=.7\columnwidth]{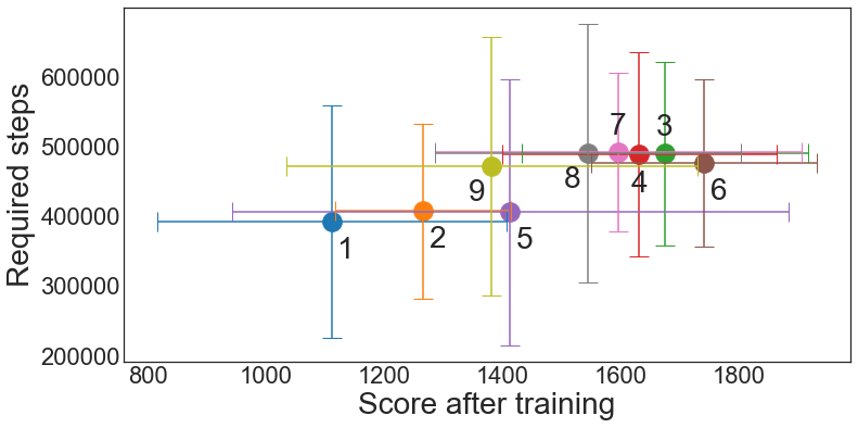}
	\vspace{-4pt}
	\caption{\label{fg:hopper_comp}Average score during the last 10\% of training steps (horizontal axis) and the average number of steps needed to reach that score for the first time (vertical axis) of all tested algorithms in the PyBullet Hopper-v0 environment. The error bars represent one standard deviation in both axes.}
\end{figure}

Hyperparameters used for training are given in the \hyperref[ap:4]{Appendix}. We have run simulations with 20 random seeds to obtain statistically meaningful results. Benchmark results are shown in Table~\ref{tb:hopper}. Furthermore, the learning curves of DDPG and the best-performing UA-DDPG are displayed in  Figure~\ref{fg:hopgraph}. To further support comparison among algorithms, we plot scores and steps until convergence in Figure~\ref{fg:hopper_comp}.
\begin{table}[h]
	\caption{\label{tb:hopper}Results for the `HopperBulletEnv-v0' environment. The best score is in bold. The errors represent one standard deviation.}
	\centering
	\setlength\tabcolsep{3pt}
	\scalebox{0.9}{
	\begin{tabular}{ccccccccc}
		\hline 
		No. & Algorithm & $N$ & $M$ & $K$ & $T_{\exp}$ & $p_{\min}$ & $\beta$ & Return
		\\\hline
		1 & DDPG &1&1&1&0&0& 1 &
		1111 $\pm$ 296
		\\
		2 & Dist-DDPG &12&1&1&0&0&$\forall\beta_i=1/12$&
		1266 $\pm$ 149
		\\
		3 & UA-DDPG &12&3&3&2e5&0.1&$\forall\beta_i=1/12$&
		1675 $\pm$ 242
		\\
		4 & UA-DDPG &12&3&1&2e5&0.1&$\forall\beta_i=1/12$&
		1632 $\pm$ 231
		\\
		5 & UA-DDPG &12&3&3&2e5&0&$\forall\beta_i=1/12$&
		1413 $\pm$ 471
		\\
		6 & UA-DDPG &12&3&3&2e5&0.1&
		$\left[\hspace{-3pt}\begin{array}{c}
		\beta_{1,...,4}=1/4 \\
		\beta_{5,...,12}=0
		\end{array}\hspace{-3pt}\right]$&
		{$\bf 1742 \pm 191$}
		\\
		7 & UA-DDPG &12&3&1&2e5&0.2&$\forall\beta_i=1/12$&
		1595 $\pm$ 310
		\\
		8 & UA-DDPG &12&3&1&0&0.1&$\forall\beta_i=1/12$&
		1544 $\pm$ 259
		\\
		9 & UA-DDPG &12&2&1&2e5&0.1&$\forall\beta_i=1/12$&
		1382 $\pm$ 348
		\\\hline 
	\end{tabular}
	}
\end{table}

On these results we can make several observations.
\begin{itemize}
\item 
UA-DDPG (No.6) outperforms DDPG significantly, with 57\% increase of return (from 1111 to 1742).
\item 
The superiority of No.2 ($N=12$) to No.1 ($N=1$) reveals that the use of distributional RL framework alone is capable of boosting the return by 14\%.
\item
No.3 ($K=3$) and No.4 ($K=1$) have a very similar performance, suggesting that the use of multiple actors does not substantially contribute to performance gain in this environment.
\item 
No.3 ($p_{\min}=0.1$) outperforms No.5 ($p_{\min}=0$) by 19\%, showing that exploratory actions are beneficial for learning not only at the beginning of training but throughout the entire training period.
\item
No.6 (risk-averse policy) is better than No.3 (risk-neutral policy) by 4\%. Although this difference is small, it is not unreasonable to speculate that avoidance of risky actions could help prevent the robot from falling down.
\item
The performance of No.4 ($p_{\min}=0.1$) and No.7 ($p_{\min}=0.2$) are very close. Hence the algorithm is robust against the choice of the exploration rate.
\item 
No.4 ($T_{\exp}=$2e5) outperforms No.8 ($T_{\exp}=0$) by 6\%. Thus, initial exploration of the environment does boost performance. However this gap is 6 times smaller than the gap between No.3 ($p_{\min}=0.1$) and No.5 ($p_{\min}=0$).
\item
No.4 ($M=3$) outperforms No.9 ($M=2$) by 18\%, implying that performance is significantly boosted through more accurate estimation of epistemic uncertainty for larger $M$. 
\end{itemize}
These insights seem to indicate that UA-DDPG’s huge performance gain over DDPG actually comprises many pieces of individually small improvement that come from all the characteristics of UA-DDPG. Although numerical data quoted above are likely to be highly specific to the current Hopper environment and not easily generalizable, one can at least conclude that $N$, $M$, $K$, $T_{\exp}$, $p_{\min}$ and $\{\beta_i\}$ must all be deliberately configured in order to derive the full potential of UA-DDPG. At present, we do not have a general guideline on the selection of these hyperparameters and leave it to future work.

We have also conducted numerical experiments in the  `HalfCheetahBulletEnv-v0' environment, the action space of which is 6 dimensional and is hence more challenging than the Hopper environment. We have confirmed that UA-DDPG outperforms DDPG significantly (results not shown due to the limitation of space).

\subsection{Power grid control}
There is a growing global trend towards decarbonization with use of renewable energy such as wind and solar energy. To realize a safe and robust society on top of volatile natural power resources, it is imperative to construct and maintain a highly optimized transmission and distribution system. There have been some works on applications of deep RL to energy grid management \cite{Foruzan2018,Sogabe2018,DBLP:journals/arc/YangZLZ20,Nakabi2021}. 

In this subsection, we apply RL to a power grid simulator PowerGym \cite{powergym2021,DBLP:journals/corr/abs-2109-08512}. There are four default environments (“13Bus”, “34Bus”, “123Bus”, and “8500Node”) in increasing order of complexity. We shall use the “34Bus” environment as a testbed for our approach. In this environment, the circuit is equipped with 2 capacitors, 6 regulators and 2 batteries. At every time step we must control these devices to minimize the sum of three losses: voltage violation loss, control error, and power loss. The action space is 10 dimensions and the state space is 107 dimensions. The actions are discrete: the first two actions associated with capacitors are binary, while the other 8 actions take integer values from 0 to 32. RL methods with continuous actions can be employed with an appropriate rounding of actions to integers. A single episode consists of 24 time steps, representing a day. (This setting can be easily changed manually but we leave it as it is.) There are 16 load profiles available, each of which consists of data of 24-hour loads at every bus. During training and testing, we randomly switch the load profile at the end of an episode. 

\begin{figure}[h]
	\centering
	\includegraphics[width=.8\columnwidth]{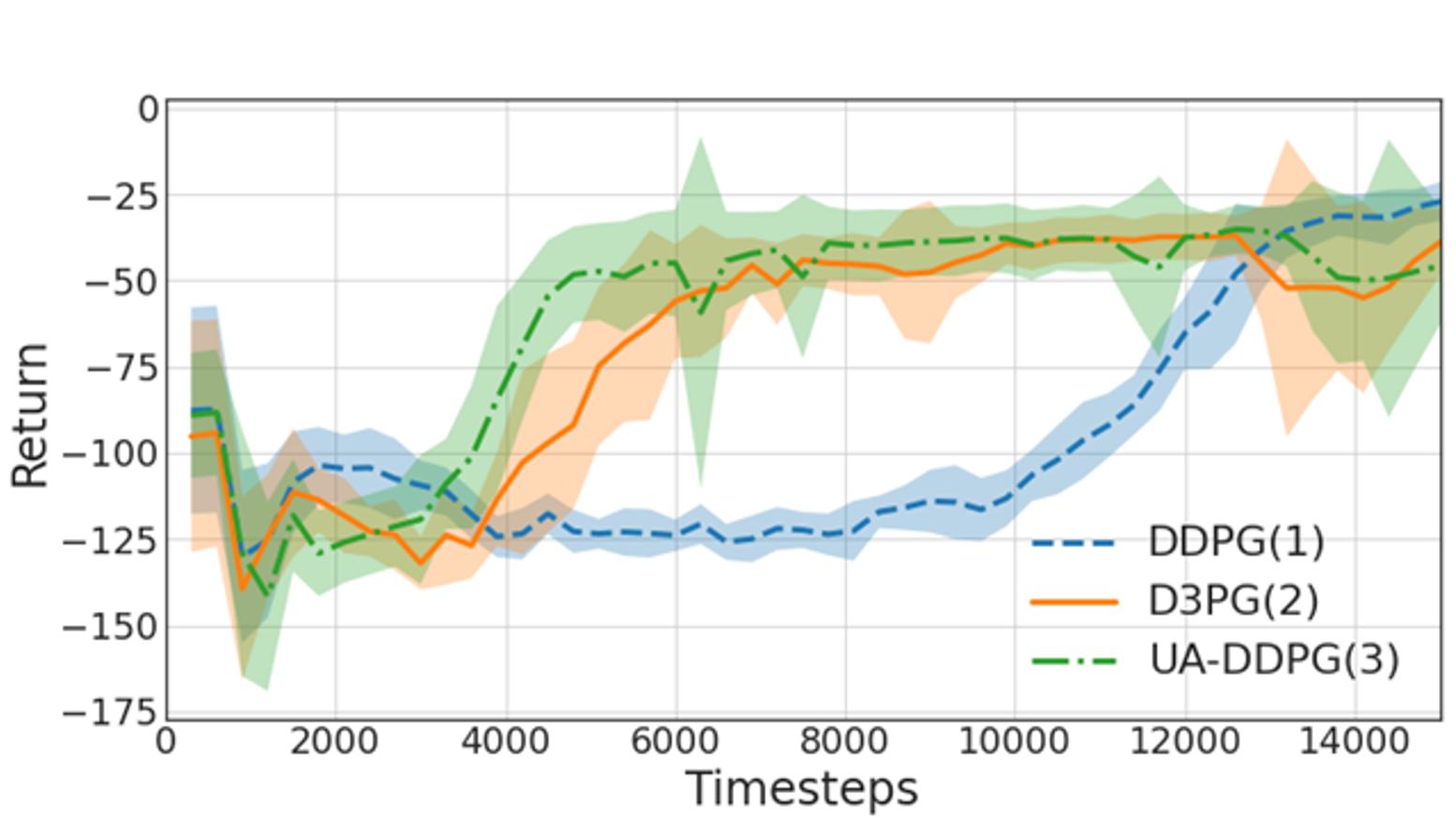}
	\vspace{-4pt}
	\caption{\label{fg:pggraph}Learning curves for the PowerGym 34Bus environment. Shaded areas represent mean ± one standard deviation.}
\end{figure}

Hyperparameters are given in the \hyperref[ap:4]{Appendix}. We have run simulations with 10 random seeds. The results of our numerical experiments are summarized in Table~\ref{tb:pg} and the learning curves are displayed in Figure~\ref{fg:pggraph}.
\begin{table}[h]
	\caption{\label{tb:pg}Results for the 34Bus environment in PowerGym. The best score is in bold. The errors represent one standard deviation.}
	\centering
	\setlength\tabcolsep{3pt}
	\scalebox{0.83}{
	\begin{tabular}{cccccccccc}
		\hline 
		No. & Algorithm & $N$ & $M$ & $K$ & $T_{\exp}$ & $p_{\min}$ & $\beta$ & $\begin{array}{c}
		\text{Return at}\\\text{step 4800}
		\end{array}$ & 
		$\begin{array}{c}
		\text{Return at}\\\text{step 15000}
		\end{array}$
		\\\hline
		1 & DDPG &1&1&1&0&0&1& $-122.8\pm6.3$ &
		{$\bf -27.1 \pm 5.6$}
		\\
		2 & Dist-DDPG &10&1&1&0&0&$\forall\beta_i=1/10$
		& $-91.9 \pm 24.5$ & $-38.9\pm 9.7$
		\\
		3 & UA-DDPG &10&3&1&1e4&0.2&$\forall\beta_i=1/10$
		& {$\bf -48.3\pm 14.0$} & $-45.8\pm17.3$
		\\\hline 
	\end{tabular}
	}
\end{table}
It is impressive that distributional DDPG (Dist-DDPG) outperforms DDPG by a large margin, only requiring half the training steps of DDPG to reach the return of $-50$. It may reflect the difficulty of DDPG to adapt to a highly stochastic environment. Note also that UA-DDPG’s learning is even faster than Dist-DDPG, which is clear from their average returns at the 4800th step in Table~\ref{tb:pg}. However, Dist-DDPG and UA-DDPG exhibit curious degradation beyond 13000 steps. Overall, we conclude that (1) UA-DDPG improves DDPG substantially (reduction of training time by ca. 60\%), (2) exploration based on epistemic uncertainty makes UA-DDPG converge faster than Dist-DDPG (reduction of training time by ca. 20\%), and (3) performances of learned policies at the end of training are similar. The last point is in contrast with the Hopper environment in the last subsection. This could be indicating that epistemic exploration in a stochastic environment like Bus34 may not be as straightforward as in a deterministic environment like Hopper. This issue merits further investigation.

\section{Conclusions and outlook\label{sc:end}}

In this work, we proposed a new generalization of DDPG, coined Uncertainty-Aware DDPG (UA-DDPG), which utilizes distributional value estimation by an ensemble of critics for efficient continuous control. UA-DDPG can take aleatoric uncertainty into accout in order to learn a risk-aware policy and, at the same time, take actions that seek high epistemic uncertainty in order to accelerate exploration. The results of our numerical experiments in various simulators seem to underscore the usefulness of UA-DDPG. 

There are several interesting directions of future research. First, it would be worthwhile to generalize UA-DDPG to state-of-the-art continuous-control algorithms such as TD3 \cite{DBLP:conf/icml/FujimotoHM18} and SAC \cite{DBLP:conf/icml/HaarnojaZAL18}. Second, while we have in this paper focused on learning of a risk-sensitive policy under a single risk criterion, it would be interesting to train the agent so that it learns policies corresponding to miscellaneous risk criteria at the same time, along the lines of \cite{Choi2021}.

\appendix[Hyperparameter setting]
\label{ap:4}
\subsection{Exploration in a cube}
{\hfil
\scalebox{0.7}{
    \begin{tabular}{cc}
        \hline
        Hyperparameter & Value
        \\\hline 
        Number of hidden layers & 2
        \\
        Layer width & 30
        \\
        Activation function & Tanh
        \\
        Batch size & 24
        \\
        Replay buffer size & 4e5
        \\
        Discount rate ($\gamma$) & 0.99
        \\
        Polyak averaging rate ($\tau$) & 0.8
        \\
        Variance of the initial network parameters ($\sigma^2$) & $1.0^2$
        \\
        Duration of exploration phase ($T_{\exp})$ & 1e5
        \\
        Minimum exploration rate ($p_{\min}$) & 0.1
        \\
        Variance of action noise $(\sigma^2_{\mathcal{A}})$ & $0.005^2$
        \\
        Environment steps for training & 4e5
        \\
        Number of random seeds & 24
        \\
        Learning rates ($\eta_A, \eta_C$) & 1e-3
        \\
        Initial random steps ($S_0$) & 5e3
        \\\hline 
    \end{tabular}
}}

\subsection{HopperBulletEnv-v0}
{\hfil 
\scalebox{0.7}{
    \begin{tabular}{cc}
        \hline
        Hyperparameter & Value
        \\\hline 
        Number of hidden layers & 2
        \\
        Layer width & 200
        \\
        Activation function & Tanh
        \\
        Batch size & 100
        \\
        Replay buffer size & 2e5
        \\
        Discount rate ($\gamma$) & 0.99
        \\
        Polyak averaging rate ($\tau$) & 0.99
        \\
        Variance of the initial network parameters ($\sigma^2$) & $1.0^2$
        \\
        Variance of action noise $(\sigma^2_{\mathcal{A}})$ & $0.1^2$
        \\
        Environment steps for training & 1e6
        \\
        Number of random seeds & 20
        \\
        Learning rate of actors ($\eta_A$) & 4e-4
        \\
        Learning rate of critics ($\eta_C$) & 8e-4
        \\
        Initial random steps ($S_0$) & 1e4
        \\\hline 
    \end{tabular}
}}

\subsection{Power grid control}
{\hfil 
\scalebox{0.7}{
    \begin{tabular}{cc}
        \hline
        Hyperparameter & Value
        \\\hline 
        Number of hidden layers & 2
        \\
        Layer width & 200
        \\
        Activation function & Tanh
        \\
        Batch size & 200
        \\
        Replay buffer size & 5e4
        \\
        Discount rate ($\gamma$) & 0.95
        \\
        Polyak averaging rate ($\tau$) & 0.95
        \\
        Variance of the initial network parameters ($\sigma^2$) & $0.1^2$
        \\
        Variance of action noise $(\sigma^2_{\mathcal{A}})$ & $0.1^2$
        \\
        Environment steps for training & 2e4
        \\
        Number of random seeds & 10
        \\
        Learning rate of actors and critics ($\eta_A, \eta_C$) & 1e-3
        \\
        Initial random steps ($S_0$) & 500
        \\\hline 
    \end{tabular}
}}

\bibliography{draft.bbl}
\end{document}